\pdfoutput=1
\documentclass[11pt]{article}

\usepackage[final]{ACL2023}

\usepackage{times}
\usepackage{latexsym}

\usepackage[T1]{fontenc}
\usepackage[utf8]{inputenc}

\usepackage{microtype}

\usepackage{inconsolata}

\usepackage{amsmath}
\usepackage{booktabs}
\usepackage[english]{babel}

\usepackage[utf8]{inputenc}

\usepackage{microtype}

\usepackage{multirow}
\usepackage{graphicx}

\usepackage{inconsolata}
\usepackage{dashrule}
\usepackage{CJKutf8}
\newcommand*\samethanks[1][\value{footnote}]{\footnotemark[#1]}

\usepackage{verbatim}

\title{Investigating Glyph-Phonetic Information for Chinese Spell Checking:
What Works and What's Next?}

\author{Xiaotian Zhang \thanks{\ \  These two authors contributed equally.}, \quad Yanjun Zheng \samethanks, \quad Hang Yan, \quad Xipeng Qiu \thanks{\ \  Corresponding author.} \\
    Shanghai Key Laboratory of Intelligent Information Processing, Fudan University \\
School of Computer Science, Fudan University \\
  {\{xiaotianzhang21, yanjunzheng21\}@m.fudan.edu.cn} \quad{\{hyan19, xpqiu\}@fudan.edu.cn}}  
\begin{document}

\maketitle

\begin{abstract}
%摘要主要说我们思考两个问题，一个预训练模型是否知道这些相似性,二是通过probe分析现有模型性能瓶颈的原因，
%
\begin{comment}
While Pre-trained Chinese Language models have achieved impressive performance on a wide range of NLP tasks, the Chinese Spell Checking (CSC) task is still not well resolved. 
Previous research has continued the method of introducing information such as glyphs and phonetics to enhance the ability to distinguish misspelled characters and achieved good performance improvements.
However, the generalization ability of the models remains poorly understood: whether these models contain the glyph-phonetic information of Chinese characters and whether this information is being fully utilized. 
In this paper, we seek to better understand the role played by the glyph-phonetic information of the misspelled characters in the CSC task, which suggests directions for improvement. 
Additionally, we propose a new setting to test the generalizability of the CSC model, which is more challenging and practical. We make all codes publicly available.
\end{comment}

While pre-trained Chinese language models have demonstrated impressive performance on a wide range of NLP tasks, the Chinese Spell Checking (CSC) task remains a challenge. 
Previous research has explored using information such as glyphs and pronunciations to improve the ability of CSC models to distinguish misspelled characters, with good results at the accuracy level on public datasets. 
However, the generalization ability of these CSC models has not been well understood: it is unclear whether they incorporate glyph-phonetic information and, if so, whether this information is fully utilized. 
In this paper, we aim to better understand the role of glyph-phonetic information in the CSC task and suggest directions for improvement. Additionally, we propose a new, more challenging, and practical setting for testing the generalizability of CSC models. Our code will be released at \href{https://github.com/piglaker/ConfusionCluster}{https://github.com/piglaker/ConfusionCluster}.

\end{abstract}

\section{Introduction}

\begin{comment}
Spell Checking (SC) aims to detect and correct spelling errors in natural human texts. With Levenshtein Distance and vocabulary, SC is relatively simple for some languages like English. But for Chinese, Chinese Spell Checking (CSC) is a challenging task because of the nature of the idiographic language. Chinese has a large vocabulary including at least 3,500 common characters which leads to huge search space and an unbalanced distribution of errors~\cite{ji2021spellbert}. A common situation is that substitutions and combinations of several characters can bring about dramatic changes in the semantics of Chinese sentences, yet such combinations of characters are still grammatically correct. The CSC task restricts the output with the requirement that the original meaning and wording must be preserved as much as possible. Figure \ref{fig:csc_example} illustrates different types of errors that lead to different target characters. Previous work attempted to introduce inductive bias to model the relationship between Chinese character glyph, phonetics, and semantics~\cite{xu2021read}.
\end{comment}

\begin{figure}[t]
    \includegraphics[width=1\columnwidth]{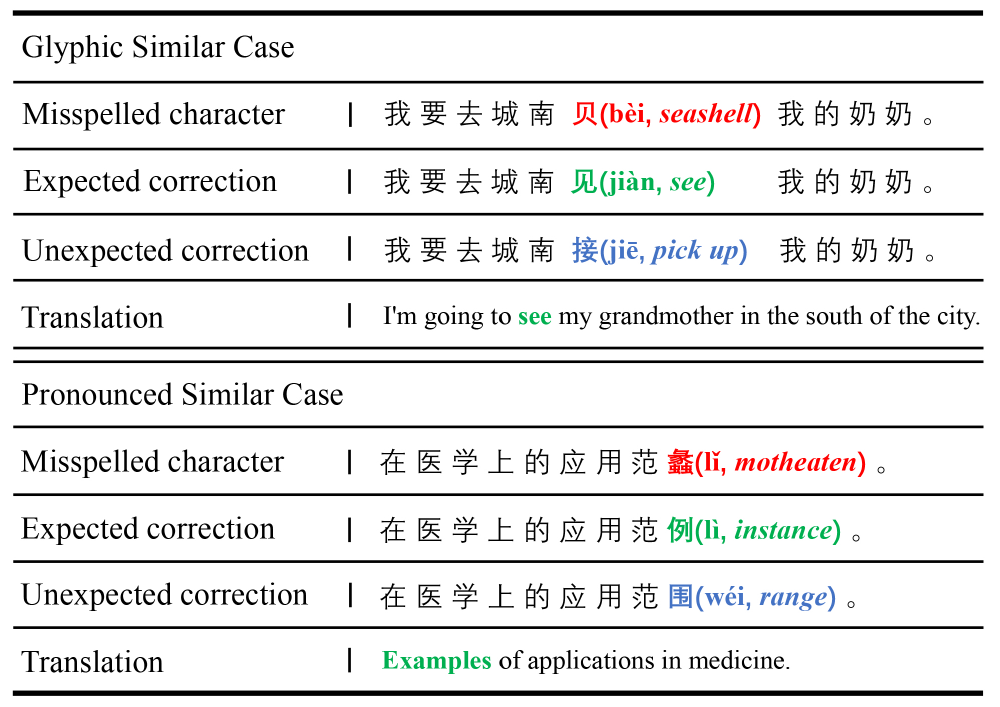}
    \caption{An example of different errors affecting CSC results. \textcolor{red}{red}/\textcolor[RGB]{0,176,80}{green}/\textcolor[RGB]{68,114,196}{blue} represents \textcolor{red}{the misspelled character}, \textcolor[RGB]{0,176,80}{the expected correction} and \textcolor[RGB]{68,114,196}{the unexpected correction.} }\label{fig:csc_example}
  \end{figure}

Spell checking (SC) is the process of detecting and correcting spelling errors in natural human texts. 
For some languages, such as English, SC is relatively straightforward, thanks to the use of tools like the Levenshtein distance and a well-defined vocabulary. 
However, for Chinese, Chinese spell checking (CSC) is a more challenging task, due to the nature of the Chinese language. 
Chinese has a large vocabulary consisting of at least 3,500 common characters, which creates a vast search space and an unbalanced distribution of errors~\cite{ji2021spellbert}. 
Moreover, substitutions or combinations of characters can significantly alter the meaning of a Chinese sentence while still being grammatically correct. 
The CSC task, therefore, requires requires the output to retain as much of the original meaning and wording as possible.
Figure \ref{fig:csc_example} shows different types of errors and corresponding target characters. 
Previous work has attempted to incorporate inductive bias to model the relationship between Chinese character glyphs, pronunciation, and semantics~\cite{xu2021read}.

\begin{CJK}{UTF8}{gbsn}
\end{CJK}

\begin{comment}
In recent years, Pre-trained Language Models (PLMs) have led to great success in a wide range of NLP tasks. With the publication of Bert~\cite{devlin2018bert}, using PLMs on CSC tasks has become a mainstream solution, such as FASpell~\cite{hong2019faspell}, Softmasked-BERT~\cite{zhang2020spelling}, SpellGCN~\cite{cheng2020spellgcn}, PLOME~\cite{liu2021plome}. Some other researchers have paid attention to the special features of Chinese characters in terms of glyphs and phonetics, trying to introduce the glyph-phonetic information and then improve the ability to distinguish misspelled characters~\cite{ji2021spellbert,liu2021plome,xu2021read}. 

However, at the same time, although the existing models have made significant improvements in performance on the CSC dataset, the generalization to real application scenarios still leaves much to be desired. How to improve the generalization capability of the CSC model? Can the current model recognize the glyph and phonetic information of the misspelled characters and utilize them to make predictions? When we rethink these previous efforts, we discovered some previously little-mentioned issues and some possible future directions:
\end{comment}

In recent years, pre-trained language models (PLMs) have shown great success in a wide range of NLP tasks. With the publication of BERT~\cite{devlin2018bert}, using PLMs for CSC tasks has become a mainstream approach, with examples including FASpell~\cite{hong2019faspell}, Softmasked-BERT~\cite{zhang2020spelling}, SpellGCN~\cite{cheng2020spellgcn}, and PLOME~\cite{liu2021plome}. Some researchers have focused on the special features of Chinese characters in terms of glyphs and pronunciations, aiming to improve the ability to distinguish misspelled characters by incorporating glyph-phonetic information~\cite{ji2021spellbert,liu2021plome,xu2021read}.
However, despite these advances, the generalization of CSC models to real-world applications remains limited. How can we improve the generalization ability of CSC models? Can current models recognize and utilize glyph-phonetic information to make predictions? As we re-examine previous work, we have identified some previously unexplored issues and potential future directions for research.

\begin{comment}
Q1: \textbf{\emph{How valuable is the misspelled character worth in correcting it?}} The misspelled characters are the misspellings of the sentences. Previous studies have emphasized the importance of the misspelled characters and we want to explore whether the CSC model must rely on the misspelled characters to do better. The specific way is that we design an experiment to explore what happens to the performance of these models if we ignore the misspelled characters and rely on the context only.
\end{comment}

Q1: \textbf{\emph{Do existing Chinese pre-trained models encode the glyph-phonetic information of Chinese characters?}} Chinese writing is morpho-semantic, and its characters contain additional semantic information. Before studying existing CSC models, we seek to investigate whether existing mainstream Chinese pre-trained language models are capable of capturing the glyph-phonetic information.

Q2: \textbf{\emph{Do existing CSC models fully utilize the glyph-phonetic information of misspelled characters to make predictions?}} Intuitively, introducing glyph-phonetic information in the CSC task can help identify misspelled characters and improve the performance of the model. However, there has been little research on whether existing CSC models effectively use glyph-phonetic information in this way. 

\begin{comment}
Q4: \textbf{\emph{What is the current bottleneck of the CSC task, dataset, or model itself? }} We note that on the mainstream dataset SIGHAN, the improvement in F1 score in the last two years is relatively small. Could this indicate that the CSC task has reached a bottleneck? What is causing this, the dataset or the model itself? 
\end{comment}

Empirically, our main observations are summarized as follows:
\begin{itemize} \setlength{\itemsep}{1pt}% 
\setlength{\parskip}{1pt}% 
\item We show that Chinese PLMs like BERT encode glyph-phonetic information without explicit introduction during pre-training, which can provide insight into the design of future Chinese pre-trained models. We also propose a simple probe task for measuring how much glyph-phonetic information is contained in a Chinese pre-trained model.
\item We analyze the ability of CSC models to exploit misspelled characters and explain why current CSC methods perform well on test sets but poorly in practice. We propose a new probe experiment and a new metric Correction with Misspelled Character Coverage Ratio (CCCR).
\item We propose a new setting for the CSC task, called isolation correction, to better test the generalizability and correction performance of CSC models. This setting alleviates the shortcuts present in the original dataset, making the CSC task more challenging and realistic. \end{itemize}

We hope that this detailed empirical study will provide follow-up researchers with more guidance on how to better incorporate glyph-phonetic information in CSC tasks and pave the way for new state-of-the-art results in this area.

\section{Related Work}
% 这里写一点相关工作，比如chinesebert，所有和字形读音有关的工作，或者和词有关的预训练比如macbert
% 
% 

\subsection{Glyph Information}

%建模图形信息的方法分为两类，一种是建立偏旁部首索引来得到字图形信息的嵌入，另一种是使用CNN之类的图像特征抽取器e2e建模

Learning glyph information from Chinese character forms has gained popularity with the rise of deep neural networks. After word embeddings~\cite{mikolov2013distributed} were proposed, early studies~\cite{sun2014radical, shi-etal-2015-radical, yin-etal-2016-multi} used radical embeddings to capture semantics, modeling graphic information by splitting characters into radicals. Another approach to modeling glyph information is to treat characters as images, using convolutional neural networks (CNNs) as glyph feature extractors~\cite{liu-etal-2010-visually, shao-etal-2017-character, dai-cai-2017-glyph, meng2019glyce}. With pre-trained language models, glyph and phonetic information are introduced end-to-end. ChineseBERT\cite{sun-etal-2021-chinesebert} is a pre-trained Chinese NLP model that flattens the image vector of input characters to obtain the glyph embedding and achieves significant performance gains across a wide range of Chinese NLP tasks.

%We consider the base model (Model Card:'junnyu/ChineseBERT-base' under Joint Laboratory of HIT and iFLYTEK Research).

\subsection{Phonetic Infomation}

Previous research has explored using phonetic information to improve natural language processing (NLP) tasks. \citeauthor{liu-etal-2019-robust} propose using both textual and phonetic information in neural machine translation (NMT) by combining them in the input embedding layer, making NMT models more robust to homophone errors. There is also work on incorporating phonetic embeddings through pre-training. \citeauthor{zhang2021correcting} propose a novel end-to-end framework for CSC with phonetic pre-training, which improves the model's ability to understand sentences with misspellings and model the similarity between characters and pinyin tokens. \citeauthor{sun-etal-2021-chinesebert} apply a CNN and max-pooling layer on the pinyin sequence to derive the pinyin embedding.

\subsection{Chinese Spell Checking}

\subsubsection{Task Description}

Under the language model framework, Chinese Spell Checking is often modeled as a conditional token prediction problem. Formally, let $X = {c_1,c_2,\ldots, c_T}$ be an input sequence with potentially misspelled characters $c_i$. The goal of this task is to discover and correct these errors by estimating the conditional probability $P(y_i|X)$ for each misspelled character $c_i$.

\subsubsection{CSC Datasets}
We conduct experiments on the benchmark SIGHAN dataset~\cite{DBLP:conf/acl-sighan/WuLL13,DBLP:conf/acl-sighan/YuLTC14,DBLP:conf/acl-sighan/TsengLCC15}, which was built from foreigners' writings and contains 3,162 texts and 461 types of errors. However, previous studies have reported poor annotation quality in SIGHAN13 and SIGHAN14~\cite{DBLP:conf/acl-sighan/WuLL13,DBLP:conf/acl-sighan/YuLTC14}, with many errors, such as the mixed usage of auxiliary characters, remaining unannotated~\cite{cheng2020spellgcn}. To address these issues and enable fair comparisons of different models, we apply our probe experiment to the entire SIGHAN dataset and use only clean SIGHAN15 for metrics in our review. The statistics of the dataset are detailed in the appendix.

\subsubsection{Methods for CSC}

To investigate the role of glyph-phonetic information in CSC, we conduct a probe experiment using different Chinese PLMs as the initial parameters of the baseline. The models we use are detailed in the appendix.
For our first probe experiment, we use the out-of-the-box BERT model as a baseline. We input the corrupted sentence into BERT and get the prediction for each token. If the predicted token for the corresponding output position is different from its input token, we consider BERT to have detected and corrected the error~\cite{zhang2022sdcl}.
We also consider two previous pre-trained methods that introduced glyph and phonetic information for CSC. PLOME~\cite{liu2021plome} is a pre-trained masked language model that jointly learns how to understand language and correct spelling errors. It masks chosen tokens with similar characters according to a confusion set and introduces phonetic prediction to learn misspelled knowledge at the phonetic level using GRU networks. RealiSe~\cite{xu2021read} leverages the multimodal information of Chinese characters by using a universal encoder for vision and a sequence modeler for pronunciations and semantics.

\subsection{Metrics}

For convenience, all Chinese Spell Checking metrics in this paper are based on the sentence level score\cite{cheng2020spellgcn}.
% only on the SIGHAN15 test set~\cite{tseng2015introduction} (Previous studies have reported that serious labeling errors on sighan13 and 14 will significantly affect the evaluation~\cite{xu2021read}).
We mix the original SIGHAN training set with the enhanced training set of 270k data generated by OCR- and ASR-based approaches~\cite{wang2018hybrid} which has been widely used in CSC task.

\section{Experiment-I: Probing for Character Glyph-Phonetic Information }
In this section, we conduct a simple MLP-based probe to explore the presence of glyph and phonetic information in Chinese PLMs and to quantify the extent to which tokens capture glyph-phonetic information. We consider glyph and phonetic information separately in this experiment.

\subsection{Glyph Probe}

\begin{figure}[htbp]
\small
\centering
\includegraphics[width=\linewidth,scale=1.00]{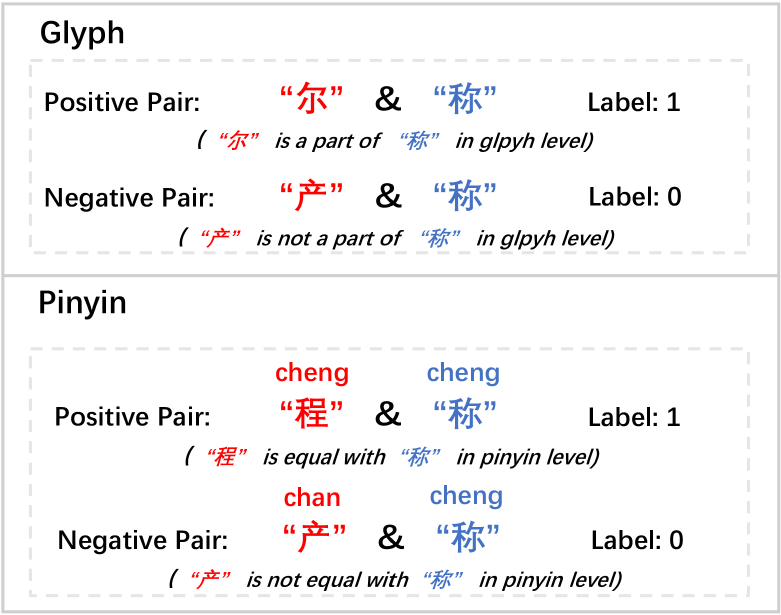}
\caption{Examples of the input and label in Experiment-I MLP Probe. We highlight the two characters in \textcolor{red}{red}/\textcolor[RGB]{68,114,196}{blue} color. }
\label{fig:mlp}
\end{figure}

For glyphs, we train a binary classifier probe to predict if one character is contained within another character. We use the frozen embeddings of these characters from Chinese PLMs as input. That is, as shown in the upper part of Figure \ref{fig:mlp}, if the probe is successful, it will predict that \begin{CJK*}{UTF8}{gbsn}“称”\end{CJK*} contains a \begin{CJK*}{UTF8}{gbsn}“尔”\end{CJK*} at the glyph level however not \begin{CJK*}{UTF8}{gbsn}“产”\end{CJK*} (it is difficult to define whether two characters are visually similar, so we use this method as a shortcut).

For the glyph probe experiment, we consider the static, non-contextualized embeddings of the following Chinese PLMs: BERT~\cite{cui2019pre}, RoBERTa~\cite{cui2019pre}, ChineseBERT~\cite{sun-etal-2021-chinesebert}, MacBERT~\cite{cui2020revisiting}, CPT~\cite{shao2021cpt}, GPT-2~\cite{radford2019language}, BART~\cite{shao2021cpt}, and T5~\cite{raffel2020exploring}. We also use Word2vec~\cite{mikolov2013efficient} as a baseline and a completely randomized Initial embedding as a control. See Appendix~\ref{PLMs_considered} for details on the models used in this experiment.

The vocabulary of different Chinese PLMs is similar. For convenience, we only consider the characters that appear in the vocabulary of BERT, and we also remove the characters that are rare and too complex in structure, such as \begin{CJK*}{UTF8}{gbsn}“圜”\end{CJK*} and \begin{CJK*}{UTF8}{gbsn}“瑷”\end{CJK*}. The statistics of our dataset for the probe are shown in Appendix \ref{probestatistic}.

We divide the character $w$ into character component $\{u_1,u_2,\ldots, u_i\}$ using a character splitting tool\footnote{https://github.com/howl-anderson/hanzi\_chaizi}. That is, \begin{CJK*}{UTF8}{gbsn}“称”\end{CJK*} will be divided into \begin{CJK*}{UTF8}{gbsn}“禾”\end{CJK*} and \begin{CJK*}{UTF8}{gbsn}“尔”\end{CJK*}. The set of all characters (e.g. \begin{CJK*}{UTF8}{gbsn}“称”\end{CJK*}) is $\mathcal{W} = \{w_1,w_2,\ldots, w_d\}$, where $d$ is number of characters. The set of all components of characters (e.g. \begin{CJK*}{UTF8}{gbsn}“禾”\end{CJK*}, \begin{CJK*}{UTF8}{gbsn}“尔”\end{CJK*}) is $\mathcal{U}=\{u_1,u_2,\ldots, u_c\}$, where $c$ is the number of components of each character. If $u_i$ exists in $w_i$, in other words, is a component of $w_i$ in glyph level, then ${u_i,w_i}$ is a positive example, and vice versa is a negative example. Then, we constructed a positive dataset ${\mathcal{D}_{pos}} = \{\{u_1,w_1\},\{u_2,w_1\},\ldots, \{u_i,w_d\}\}$, where the $u$ corresponds to $w$ separately. Also, we constructed a balanced negative dataset $\mathcal{D}_{neg} = \{\{u_1^n,w_1\},\{u_2^n,w_1\},\ldots, \{u_i^n,w_d\}\}$, where d is equal to $\mathcal{D}_{pos}$ and $u^n$ is randomly selected in the set $U$. We mix $\mathcal{D}_{pos}$ and $\mathcal{D}_{neg}$ and split the dataset into training and test according to the ratio of 80:20 to ensure that a character only appears on one side.

We train the probe on these PLMs' static non-trainable embeddings. For every ${u_i,w_i}$, we take the embedding of $u_i$ and $w_i$, and concatenation them as the input $x_i$. The classifier trains an $MLP$ to predict logits $\hat{y_i}$, which is defined as :
\begin{align}
    \hat{y_i} = \textrm{Sigmoid}(\textrm{MLP}(x_i)) \nonumber
\end{align}

To control the variables as much as possible and mitigate the effects of other factors on the probe experiment, we also experimented with the number of layers of $MLP$. The results of this are detailed in Appendix~\ref{difflayers}.

\subsection{Phonetic Probe}
For phonetics, we train another binary classifier probe to predict if two characters have the similar pronunciation, also using the frozen embeddings of these characters from Chinese PLMs as input. The meaning of 'similar' here is that the pinyin is exactly the same, but the tones can be different. That is, as shown in the lower part of Figure \ref{fig:mlp}, if the probe is successful, it will predict that \begin{CJK*}{UTF8}{gbsn}“称”\end{CJK*}\emph{(cheng)} has the similar pronunciation with \begin{CJK*}{UTF8}{gbsn}“程”\end{CJK*}\emph{(cheng)} however not \begin{CJK*}{UTF8}{gbsn}“产”\end{CJK*}\emph{(chan)}.

We consider static non-contextualized embedding of Chinese PLMs, which are the same as the glyph probe. We also mainly analyze the characters in the vocabulary of BERT, and mainly consider common characters.

The dataset construction is also similar to the glyph probe. To create positive examples, for each character $w_i$ in character list $W$, we find a character $u_i$ which has the similar pronunciation as $w_i$, then ${u_i,w_i}$ is a positive example. For each positive, we also find a character $s_i$ which has a different pronunciation from $w_i$ to construct negative example ${s_i,w_i}$. For example, the positive example is the two characters with similar pronunciation, such as \begin{CJK*}{UTF8}{gbsn}“称”\end{CJK*} \emph{(cheng)} and \begin{CJK*}{UTF8}{gbsn}“程”\end{CJK*}\emph{(cheng)}. And the negative example is the two characters with different pronunciation, such as \begin{CJK*}{UTF8}{gbsn}“称”\end{CJK*}\emph{(cheng)} and \begin{CJK*}{UTF8}{gbsn}“产”\end{CJK*}\emph{(chan)}. The divide ratio and other settings are the same as the glyph probe.

We train the probe on these PLMs' static non-trainable embeddings as the glyph probe and also concatenate the embeddings of the pairs as input.

\subsection{Results and Analysis}

\begin{figure*}[t]
    \centering
    \includegraphics[width=0.86\textwidth]{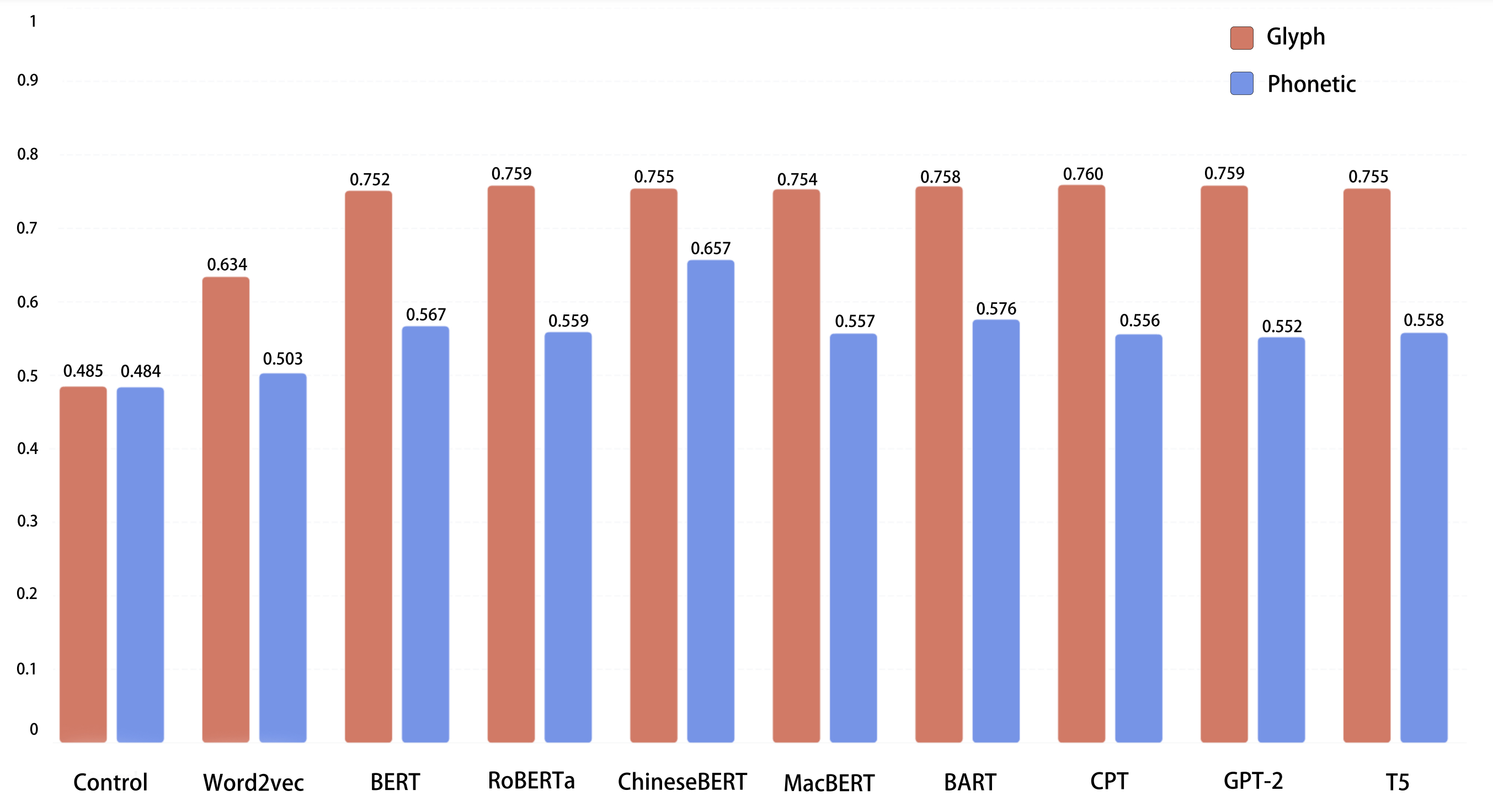}
    \caption{Results of Probe for Chinese PLMs. We found that the language models modeled by different paradigms are roughly close in perceiving graphical information but weak in speech. It is worth noting that ChineseBERT performs more significantly on this probe, probably because it explicitly introduces graphical and pronunciation information from the embedding stage.}\label{fig:1}
\end{figure*}

The following conclusions can be drawn from Figure \ref{fig:1}.

\begin{comment}
\begin{table}[htbp]
\centering
\begin{tabular}{ccc}
   \toprule
   Method & Acc. \\
   \midrule
   Control  & 0.485 \\
   Word2vec & 0.634 \\
   BERT     & 0.752 \\
   RoBERTa  & 0.759 \\
   ChineseBERT & 0.755 \\  
   MacBERT & 0.754 \\
   BART & 0.758 \\
   CPT & 0.760 \\
   GPT-2  & 0.759 \\
   T5 & 0.755 \\
   \bottomrule
\end{tabular}%

\caption{Results of Glyph Probe.}
\label{glyph_results}
\end{table}
\end{comment}

%Table \ref{glyph_results} and Table \ref{phonetic_results} show the results averaged across 5 train-test splits and different seeds, reporting on the Accuracy metric on the dataset. 

\paragraph{The Chinese PLMs encoded the glyph information of characters}
From the results, we can see that for glyphs, all models outperform the control model. 
The results of the control are close to 50\% that there is no glyph information encoded in the input embedding, and the model guesses the result randomly. 
Comparing Word2vec and other Chinese PLMs side-by-side, we find that the large-scale pre-trained model has a significant advantage over Word2vec, suggesting that large-scale pre-training can lead to better representation of characters. 
In addition, we find that the results of these Chinese PLMs are concentrated in a small interval. 
ChineseBERT boasts of introducing glyph-phonetic information, which do not have advantages in glyph.

\begin{comment}
\begin{table}[htbp]
\centering
\begin{tabular}{cc}
   \toprule
   Method  & Acc. \\
   \midrule
   Control  & 0.484 \\
   Word2vec    & 0.503 \\
   BERT         & 0.567 \\
   RoBERTa     & 0.559 \\
   ChineseBERT  & 0.657 \\
   MacBERT  & 0.557 \\
   BART  & 0.576 \\
   CPT  & 0.556 \\
   GPT-2  & 0.552 \\
   T5  & 0.558 \\
   \bottomrule
\end{tabular}%
\caption{Results of Phonetic Probe.}
\label{phonetic_results}
\end{table}
\end{comment}

\paragraph{PLMs can hardly distinguish the phonetic features of Chinese characters}

In our experiments, the control group performed similarly to the phonetic probe, with an accuracy of approximately 50\%. Unlike the glyph probe, the accuracy of Word2vec and other Chinese PLMs are also low in this probe. However, the introduction of phonetic embedding allowed ChineseBERT to perform significantly better than the other models. Our analysis suggests that current Chinese PLMs may have limited phonetic information.
%analysis here

\begin{table}[htb]
\centering
\begin{tabular}{ccc}
   \toprule
   Method & Acc. \\
   \midrule
   Control & 0.485 \\
   Word2vec  & 0.634 \\
   \midrule
   BERT    & 0.752 \\
   RoBERTa  & 0.759 \\
   ChineseBERT & 0.755 \\
   \midrule
   BERT-trained      & 0.756 \\
   RoBERTa-trained   & 0.757 \\
   ChineseBERT-trained & 0.759 \\
   \bottomrule
\end{tabular}%
\caption{Results of Probe for Models trained on the CSC task.We find that training on spell checking dataset does not enhance the graphical perception capability of models.}
\label{compare_after_csc}
\end{table}

\paragraph{Model training on the CSC task does not enrich glyph and phonetic information} 

We perform the same two probes using models fine-tuned on the SIGHAN dataset. We aim to investigate whether the training for the CSC task could add glyph and phonetic information to the embeddings, and the results are shown in Table \ref{compare_after_csc}. We found that the difference between the fine-tuned and untrained models is almost negligible, indicating that the relevant information is primarily encoded during the pre-training stage.

\section{Experiment-II: Probing for Homonym Correction}

In this experiment, we aim to explore the extent to which existing models can make use of the information from misspelled characters. To do this, we propose a new probe called Correction with Misspelled Character Coverage Ratio(CCCR), which investigates whether the model can adjust its prediction probability distribution based on the glyph-phonetic information of misspelled characters when making predictions.

\subsection{Correction with Misspelled Character Coverage Ratio}

\paragraph{Measure models utilizing the misspelled characters} 
In this paper, we propose a method to evaluate the ability of a model to make predictions using additional information from misspelled characters, as well as to assess whether the model contains glyph-phonetic information.

\begin{CJK}{UTF8}{gbsn}
\begin{figure}[t]
\resizebox{\linewidth}{!}{%
\includegraphics[width=1\linewidth]{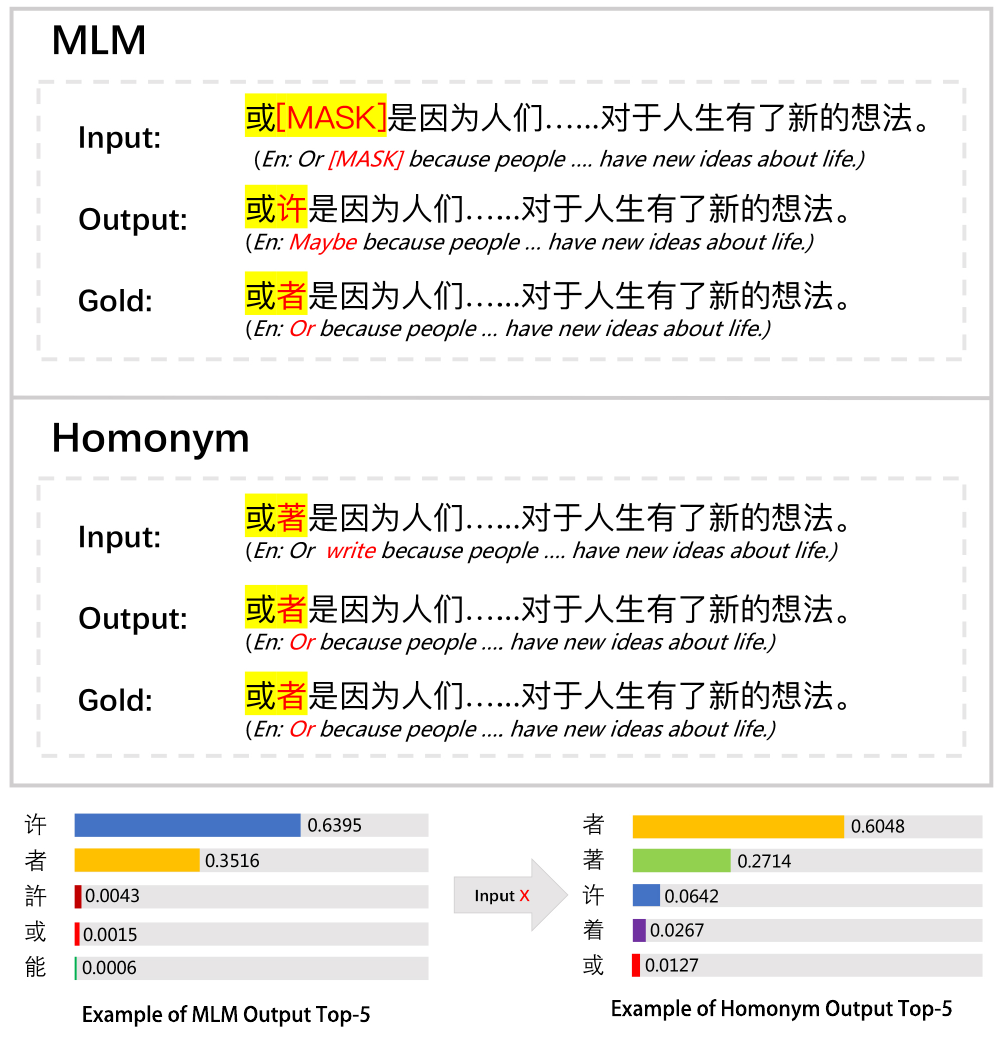}}
\caption{Take BERT as an example. The first half shows examples of MLM and Homonym respectively. The bottom half shows the change in the probability distribution predicted by the model in this example. }\label{fig:cccr_example}
\end{figure}
\end{CJK}

Assume that $\mathcal{C}$ is a combination set of all possible finite-length sentence $C_i$ in the languages $L$,  $\mathcal{C}=\left\{C_{0},...,C_{i},...\right\}$, $C_i=\left\{c_{i,1},...,c_{i,n},... \right\}$, while $c_{i,j} \in L$. Let sentence $C_{i}^{n,a}$ be $C_{i}^{n, a}=\left\{c_{i,1},...,c_{i, n-1},a,c_{i, n+1},...\right\}$, then assume that the representation learning model, let $H^{w}(C)$ be the hiddens of model $w$, $X_i$ is an example in $\mathcal{C}$, For model $w$, the probability of token in position $i$ should be:

\begin{align}
    P\left(y_i=j|X_i, w\right)=\textrm{softmax}\left(WH^{w}(X_i)+b\right)[j]
\nonumber
\label{softmax}
\end{align}

Dataset $\mathcal{D}$ is a subset of $\mathcal{C}$, Then we can approximate the probability of the model. The CCCR is composed of $\mathcal{MLM}$ and $Homonym$. The former indicates which samples need the information on misspelled characters to be corrected while the latter shows which sample models adjust the output distribution. We take the intersection to get the frequency of whether the model is adjusted for the samples whose distribution should be adjusted.

\begin{table*}[]
\normalsize
\centering

\begin{tabular}{cccccccc}
\toprule
Method & MLM & Homonym & CCCR & Precision & Recall & F1 \\
\midrule
Baseline & - & -  & 15.61 & - & - & - \\
BERT-Initial & 45.58 & 64.87 & 34.57 & - & - & - \\
RoBERTa-Initial & 46.53 & 60.19 & 28.17 & - & - & - \\
ChineseBERT-Initial & 44.97 & 62.22 & 31.17 & - & - & - \\
\midrule
BERT & 48.57 & 67.73 & 41.67 & 43.72 & 26.93 & 33.32 \\
RoBERTa & 48.70 & 64.80 & 36.12 & 39.82 & 27.14 & 32.27 \\
ChineseBERT & 46.33 & 67.39 & 40.32 & 42.56 & 27.26 & 33.23 \\
\midrule
PLOME & 55.63 & 88.38 & 80.83 & 42.63 & 37.15 & 39.70 \\
ReaLiSe & 51.29 & 84.23 & 78.14 & 52.26 & 19.23 & 28.11 \\
\bottomrule
\end{tabular}%

\caption{ Model performance in the isolation correction setting of SIGHAN15. '-Initial' means without any training. }
\label{tab:table isolation results}
\end{table*}

\paragraph{$\mathcal{MLM}$} MLM is a subset of dataset $\mathcal{D}$. For all input sentence $C_i \in \mathcal{D},  C_i = \{c_1,c_2,[MASK],\ldots, c_T\}$ and the position of $[MASK]$ is spelling error, let special token $mask=[MASK]$ , $C_i \in MLM$ if:

\begin{equation}
\setlength\abovedisplayskip{-2pt}%shrink space
\setlength\belowdisplayskip{-4pt}
\resizebox{\columnwidth}{!}{
$P\left(y_i=noise\middle|C_{i}^{n,mask},w\right)>P\left(y_i=Y_i\middle|C_i^{n,mask},w\right)$
}
\nonumber
\end{equation}

\paragraph{$Homonym$} Same to MLM, For input sentence $C_i \in \mathcal{D}, C_i = \{c_1,c_2,c_{misspelled},\ldots, c_T\}$ and the position of $c_{misspelled}$ is spelling error. For all sentences $C_i$ in the dataset $\mathcal{D}$, $C_i \in Homonym$ if:

\begin{equation}
\setlength\abovedisplayskip{-2pt}%shrink space
\setlength\belowdisplayskip{-4pt}
\resizebox{\columnwidth}{!}{
$P(y_i=Y_i|C_{i}^{n,c_{misspelled}}, w))>P(y_i=noise|C_{i}^{n,c_{misspelled}},w)$
}
\nonumber
\end{equation}

\paragraph{Correction with Misspelled Character Coverage Ratio (CCCR)} The measured ratio is used to describe the lower bound of the probability that the model uses the information of the misspelled characters for the sentences $C_i$ in the dataset $\mathcal{C}$.

\begin{equation}
\setlength\abovedisplayskip{-2pt}%shrink space
\setlength\belowdisplayskip{-4pt}
\resizebox{\columnwidth}{!}{
$CCCR = \frac{|\{{C_i \mid C_i \in MLM} \land {C_i \in Homonym} \}|} { | \{C_i \mid C_i \in MLM \} | }$
}
\nonumber
\end{equation}

\paragraph{Baseline} Independently, we give an estimation method for the base value. Given model $w$, 
$noise$, dataset $\mathcal{D}$, ground truth correction $y$. The baseline of CCCR should be estimated as:

\begin{align}
guess_i = \frac{P(y_i=noise|C_{i}^{n, mask},w)}{1 - P( y_i=noise| C_{i}^{n,mask},w)}
\nonumber
\end{align}

\begin{align}
CCCR_{baseline} = \frac{\sum_{i \in S} \left\{ 1 *  guess_i\right\}}{
 | \left\{ C_i \mid C_i \in MLM \right\} |
}
\nonumber
\end{align}

The baseline can be understood as a model with no glyph-phonetic information at all, and the probability of being able to guess the correct answer. But no such language model exists. For this purpose,  instead of inputting the misspelled characters into the model, we artificially design strategies for the model to randomly guess answers by weight from the remaining candidates, which is equivalent to the probability of being able to guess correctly.

This probability is comparable to CCCR. CCCR restricts the condition for $y$ to overtake $noise$. In the case of baseline, considering rearranging the candidates, the probability of $y$ overtaking noise can also be re-normalized by probability.

\begin{table}[htb]
\normalsize
\centering
\resizebox{\columnwidth}{!}{%
\begin{tabular}{cccccccc}
\toprule
Method & MLM & Homonym & CCCR & Precision & Recall & F1 \\
\midrule
Baseline & - & -  & 15.61 & - & - & - \\
BERT & 52.64 & 95.78 & 92.1 & 70.15 & 75.46 & 72.71 \\
RoBERTa& 47.07 & 95.92 & 91.77 & 70.49 &74.91 & 72.63 \\
ChineseBERT& 48.57 & 97.62 & 96.83 & 73.24 & 76.75 & 74.59 \\
\bottomrule
\end{tabular}%
}
\caption{ Model performance in the original version of SIGHAN15, which is finetuned. We found that the CCCR of the model fine-tuned on the CSC dataset is very high. We found that this is caused by overlapped pairs in the training data.}
\label{tab:table Correction Setting}
\end{table}

\subsection{Isolation Correction Setting Experiment}

\begin{table}
\normalsize
\centering
\begin{tabular}{@{}lrr@{}}
\toprule
 & \#Pairs Count & \#sent \\ \midrule
Training Set & 23140 &284196\\
Test Set & 824 & 2162\\ \midrule
Training Set $\cap$ Test Set & 799 &-\\
Training Set $\cup$ Test Set & 23165 &-\\ \midrule 
Isolation Training Set & 20758 &230525\\ 
Isolation Test Set & 824 & 2162\\\bottomrule
\end{tabular}

\caption{The overlap of the correction pairs in the train and test sets and the statistics of the isolation SIGHAN set.}
\label{statistic11}
\end{table}

In the previous section, we test CCCR on the model finetuned on the SIGHAN dataset then found the CCCR of the models approached 92\%. The results are shown in Table \ref{tab:table Correction Setting}.
%Appendix \ref{CCCR_finetuned}. 
As shown in Table \ref{statistic11}, we analyze the overlap of correction pairs in the training and test sets in the SIGHAN dataset. 

To test the model generalization ability, we design Isolation Correction Task, which removes all overlapped pairs in the training set and duplicate pairs in the test set. With isolation, the training set is reduced by about 16\%. We believe that such a setup can better test the generalizability of the model and is more challenging and practical. Within the CCCR probe, We explore the ability of the model whether rely on its information, not just the ability to remember the content on the isolated SIGHAN dataset. The result is shown in Table \ref{tab:table isolation results}

Between CCCR and F1 score, the mismatch phenomenon we refer to as stereotype is observed. The correction pair remembered while training harms the generalization of models.

\begin{figure}[htb]
    \centering
    \includegraphics[width=1\columnwidth]{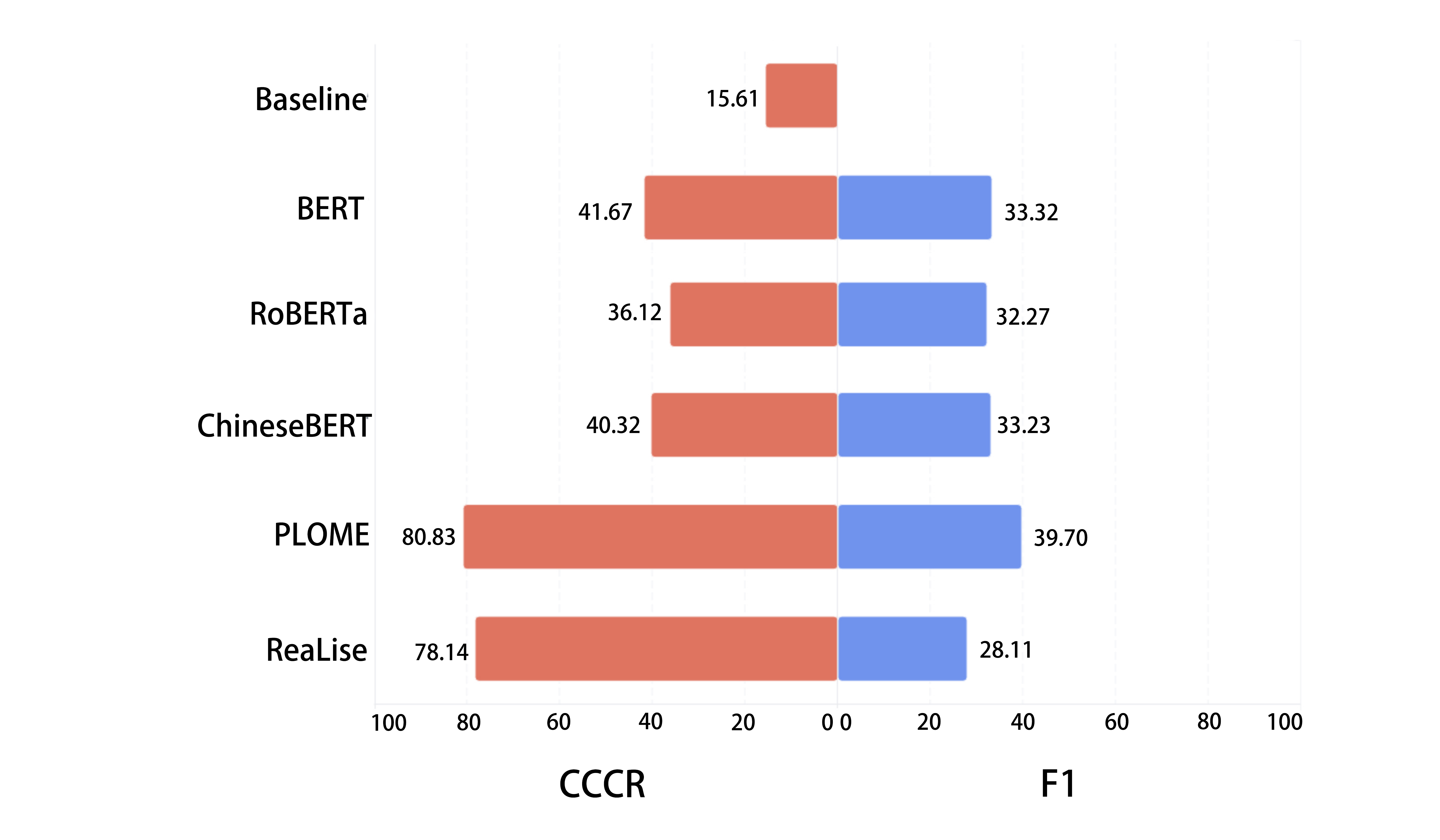}
    \caption{Results of CCCR Probe. We observe CCCR and F1 values mismatch. and for the pre-trained CSC model, we observe a phenomenon we call stereotype, which maintains a high CCCR under the isolation setting while performing worse on the F1 score, implying that stereotyping during pre-training weakens the generalization of the model.}\label{fig:2}
\end{figure}

\begin{comment}
\begin{table}[]
\small
\centering
\resizebox{\columnwidth}{!}{%
\begin{tabular}{ccccc}
\toprule
Method & MLM   & Homonym & Inter. & CCCR     \\
\midrule
Baseline    & - & -   & -  & 16.30 \\
BERT        & 45.58 & 64.87   & 15.76  & 34.57 \\
RoBERTa     & 46.53 & 60.19   & 13.11  & 28.17 \\
ChineseBERT & 44.97 & 62.22   & 14.02  & 31.17 \\
\bottomrule
\end{tabular}%
}
\caption{Initial.}
\label{tab:my-table6}
\end{table}
\end{comment}

\subsection{Results and Analysis}
We conducted experiments on three generic Chinese PLMs, BERT, RoBERTa, and ChineseBERT, and two CSC Models, PLOME, and Realise. We compare the metrics difference between the Initial model and the model after finetuning the isolation training set. The result is shown in Table~\ref{tab:table isolation results}.

\noindent \textbf{CCCR and F1 values mismatch} Our experimental results show that the CCCR and F1 values mismatch for CSC models. In the isolation training setting, we observed that the F1 values of PLOME and ReaLise are both significantly lower than their performance in Table \ref{tab:table isolation results}, indicating that their ability to make correct predictions is primarily based on the memory of correction pairs in the training set. However, their CCCR values remained high, suggesting that they are able to discriminate glyph-phonetic information but are not able to correct it effectively.
%Therefore, it is better to find ways to make better use of glyph and phonetic information in future CSC than to continue to introduce them in the model.

\noindent \textbf{Stereotype harm the generalization ability of the model in Isolation Correction Experiments} These results suggest that the correction performance of the models is primarily dependent on their memory ability and that a strong reliance on memory can hinder generalization. The poor performance in the isolation setting indicates that none of the current methods generalize well, which presents a significant challenge for future CSC research. We recommend that future research in this field follow the isolation experiment setting to address this challenge.

\section{Conclusion}
\label{sec:bibtex}

In this paper, we have explored the role of glyph-phonetic information from misspelled characters in Chinese Spell Checking (CSC). Based on our experimental results, we have reached the following conclusions:
\begin{itemize}
\setlength{\itemsep}{1pt}%
\setlength{\parskip}{1pt}%
  \item Current Chinese PLMs encoded some glyph information, but little phonetic information.
  \item Existing CSC models could not fully utilize the glyph-phonetic information of misspelled characters to make predictions. 
  \item There is a large amount of overlap between the training and test sets of SIGHAN dataset, which is not conducive to testing the generalizability of the CSC model. We propose a more challenging and practical setting to test the generalizability of the CSC task.
\end{itemize}

Our detailed observations can provide valuable insights for future research in this field. It is clear that a more explicit treatment of glyph-phonetic information is necessary, and researchers should consider how to fully utilize this information to improve the generalizability of their CSC models. We welcome follow-up researchers to verify the generalizability of their models using our proposed new setting.

\begin{comment}
\noindent \textbf{\emph{What is the current bottleneck of the Chinese error correction task, dataset, or model itself? }} We believe that there are problems with the mainstream dataset SIGHAN, such as some labeling errors and too much overlap between the training and test sets. Thus we propose a new, more practical setting based on the SIGHAN dataset. With this new benchmark, the generalization and extrapolation of the model will be more tested, which is more valuable and challenging in practice. With this setup, there is still plenty of room for the model to progress. We will release this benchmark later.
\end{comment}

\section{Limitation}

\subsection{Limited number of CSC models tested} During our research, we encountered difficulties in reproducing previous models due to unmaintained open source projects or the inability to reproduce the results claimed in the papers. As a result, we are unable to test all of the available models.

\subsection{Limited datasets for evaluating model performance} There are currently few datasets available for the CSC task, and the mainstream SIGHAN dataset is relatively small. The limited size of the data used to calculate the metrics may not accurately reflect the performance of the models. Furthermore, we found that the quality of the test set is poor, the field is narrow, and there is a large gap between the test set and real-world scenarios.

\section*{Acknowledgments} 
This work was supported by the National Key Research and Development Program of China (No.2020AAA0106700) and National Natural Science Foundation of China (No.62022027). We would like to express our gratitude to all the reviewers for their diligent, careful, and responsible feedback.

\bibliography{custom}
\bibliographystyle{acl_natbib}

%\clearpage
\appendix

\section{The Statistic of SIGHAN Dataset}
\label{statistic}
\begin{table}[htbp]
\begin{tabular}{@{}lrrr@{}}
\toprule
Training Set & \#Sent & Avg. Length & \#Errors \\ \midrule
SIGHAN14 & 3,437 & 49.6 & 5,122 \\
SIGHAN15 & 2,338 & 31.3 & 3,037  \\ 
Wang271K & 271,329 & 42.6 & 381,962  \\ \midrule
Total & 277,104 & 42.6 & 390,121 \\ \midrule \midrule
Test Set & \#Sent & Avg. Length & \#Errors \\ \midrule
SIGHAN14 & 1,062 & 50.0 & 771 \\
SIGHAN15 & 1,100 & 30.6 & 703 \\ \midrule
Total & 2,162 & 40.5 & 1,474 \\ \bottomrule
\end{tabular}
\label{tab:statistic}%
\caption{Statistics of the SIGHAN datasets.}
\end{table}

\section{The Experimental Results of Different Parameters}
\label{groups of para}

\begin{table}[htbp]
\centering
\resizebox{\columnwidth}{!}{%
\begin{tabular}{@{}llccccccccc@{}}
\toprule
\multicolumn{2}{l}{\multirow{2}{*}{}}                                             & \multicolumn{3}{c}{Para 1} & \multicolumn{3}{c}{Para 2} & \multicolumn{3}{c}{Para 3} \\ \cmidrule(l){3-11} 
\multicolumn{2}{c}{} &
  \multicolumn{1}{c}{Precision} &
  \multicolumn{1}{c}{Recall} &
  \multicolumn{1}{c}{F1} &
  \multicolumn{1}{c}{Precision} &
  \multicolumn{1}{c}{Recall} &
  \multicolumn{1}{c}{F1} &
  \multicolumn{1}{c}{Precision} &
  \multicolumn{1}{c}{Recall} &
  \multicolumn{1}{c}{F1} \\ \midrule
\multicolumn{1}{l|}{\multirow{3}{*}{SIGHAN14}} & \multicolumn{1}{l|}{BERT}        & 65.7    & 68.7    & 67.2   & 65.3    & 70.1    & 67.6   & 60.2    & 63.7    & 61.9   \\
\multicolumn{1}{l|}{}                          & \multicolumn{1}{l|}{RoBERTa}     & 64.9    & 69.3    & 67.1   & 64.0    & 67.6    & 65.7   & 58.8    & 64.9    & 62.7   \\ 
\multicolumn{1}{l|}{}                          & \multicolumn{1}{l|}{ChineseBERT} & 63.5    & 68.2    & 65.7   & 62.1    & 66.6    & 64.3   & 65.5    & 70.3    & 67.8   \\ \cmidrule(r){1-2}
\multicolumn{1}{l|}{\multirow{3}{*}{SIGHAN15}} & \multicolumn{1}{l|}{BERT}        & 74.1    & 78.4    & 76.2   & 71.8    & 76.9    & 74.3   & 70.1    & 72.6    & 71.3   \\
\multicolumn{1}{l|}{}                          & \multicolumn{1}{l|}{RoBERTa}     & 73.9    & 78.0    & 75.9   & 71.9    & 76.0    & 74.9   & 68.0    & 73.8    & 70.7   \\ 
\multicolumn{1}{l|}{}                          & \multicolumn{1}{l|}{ChineseBERT} & 73.3    & 78.5    & 75.8   & 72.4    & 77.4    & 74.8   & 73.2    & 76.7    & 74.9   \\ \bottomrule
\end{tabular}%
}
\caption{All results for fine-tuning pre-trained models  in raw data. }
\label{tab:my-table}
\end{table}

%\subsection{Training Details}\label{train para}

In Experiment I, we use the average of three sets of training parameters as the final result, which is due to the large fluctuation of performance on the test set during the experiment.

We use the pre-trained weight realized by \cite{cui2020revisiting}. For all of our models, we use the AdamW optimizer \cite{DBLP:conf/iclr/LoshchilovH19} to optimize our model for 20 epochs, the learning rate is set to be 5e-5, the batch size is 48 and the warm-up ratio is set to be 0.3.

\section{Probe details}\label{probe_details}
Our implementation uses PyTorch\cite{paszke2019pytorch} and HuggingFace\cite{wolf2020transformers}. The probes for each MLP are trained separately starting with random initialization weights. We train the probe via a binary classification task, using the Adam optimizer and Cross Entropy Loss.

\subsection{PLMs considered}\label{PLMs_considered}

We selected several mainstream Chinese PLMs as our research objects, along with their model card on Huggingface:

\textbf{BERT-Chinese}~\cite{cui2019pre} consists of two pre-training tasks: Masked Language Model (MLM) and Next Sentence Prediction (NSP), and introducing a strategy called whole word masking (wwm) for optimizing the original masking in the MLM task.
We consider the base model with 110 Million parameters. Model Card:'hfl/chinese-bert-wwm-ext' under Joint Laboratory of HIT and iFLYTEK Research.

\textbf{RoBERTa-Chinese}~\cite{cui2019pre} removes the next sentence prediction task and uses dynamic masking in the MLM task.
We also consider the base model. Model Card:'hfl/chinese-roberta-wwm-ext' under Joint Laboratory of HIT and iFLYTEK Research.

\textbf{ChineseBERT}~\cite{sun-etal-2021-chinesebert} proposes to integrate the glyph-phonetic information of Chinese characters into the Chinese pre-training model to enhance the ability to model the Chinese corpus. We consider the base model. Model Card:'junnyu/ChineseBERT-base' under Joint Laboratory of HIT and iFLYTEK Research.

\textbf{MacBERT}~\cite{cui2020revisiting} suggests that $[MASK]$ token should not be used for masking, but similar words should be used for masking because $[MASK]$ has rarely appeared in the fine-tuning phase. We also consider the base model. Model Card:'hfl/chinese-macbert-base' under Joint Laboratory of HIT and iFLYTEK Research.

\textbf{CPT}~\cite{shao2021cpt} proposes a pre-trained model that takes into account both understanding and generation. Adopting a single-input multiple-output structure, allows CPT to be used flexibly in separation or combination for different downstream tasks to fully utilize the model potential. We consider the base model. Model Card:'fnlp/cpt-base' under Fudan NLP.

\textbf{BART-Chinese}~\cite{lewis2019bart,shao2021cpt} proposes a pre-training model that combines bidirectional and autoregressive approaches. BART first uses arbitrary noise to corrupt the original text and then learns the model to reconstruct the original text. In this way, BART not only handles the text generation task well but also performs well on the comprehension task. We consider the base model. Model Card:'fnlp/bart-base-chinese' under Fudan NLP.

\textbf{T5-Chinese}~\cite{raffel2020exploring,zhao2019uer} leverages a unified text-to-text format that treats various NLP tasks as Text-to-Text tasks, i.e., tasks with Text as input and Text as output, which attains state-of-the-art results on a wide variety of NLP tasks. We consider the base model. Model Card:'uer/t5-base-chinese-cluecorpussmall' under UER.

%\textbf{GPT2-Chinese}~\cite{radford2019language,zhao2019uer} uses generalized pre-training to improve natural language comprehension. The objective function is to predict the k+1 word from the first k words, and the task is much harder than completing fill-in-the-blank (predicting the middle word being masked based on contextual information). Model Card:'uer/gpt2-chinese-cluecorpussmall' under UER.

\subsection{The Statistics of Probe Dataset}\label{probestatistic}

\begin{comment}
\begin{table}[htbp]
\centering
\begin{tabular}{@{}lr@{}}
\toprule
& \#Characters Count  \\ \midrule
Total & 10432  \\  \bottomrule
\end{tabular}
\caption{The Statistics of Characters for Glyph and Phonetic Probe. }
\label{}
\end{table}
\end{comment}

\begin{table}[htbp]
\centering
\begin{tabular}{@{}lrrr@{}}
\toprule
& \#Pos. & \#Neg. & \#Total \\ \midrule
Training Set & 7968 & 7968 & 15936 \\
Test Set & 1992 & 1992 & 3984 \\  \bottomrule
\end{tabular}

\caption{The statistics of the dataset for the glyph probe.}
\label{}
\end{table}

\begin{table}[htbp]
\centering
\begin{tabular}{@{}lrrr@{}}
\toprule
& \#Pos. & \#Neg. & \#Total \\ \midrule
Training Set & 8345 & 8345 & 16690 \\
Test Set & 2087 & 2087 & 4174 \\  \bottomrule
\end{tabular}

\caption{The statistics of the dataset for the phonetic probe. }
\label{}
\end{table}

\subsection{Probing Results from Models with Different Numbers of MLP Layers}\label{difflayers}

From the experimental results, it can be seen that the number of layers of MLP has little effect on the results, and most of the results of the pre-training models are finally concentrated in the interval of 0.75-0.76. The Chinese pre-training models of the BERT family are slightly less effective when the number of layers is relatively small and similar to other Chinese pre-training models after more than three layers.

\begin{figure}[htbp]
\resizebox{\linewidth}{!}{%
\includegraphics[width=0.8\linewidth]{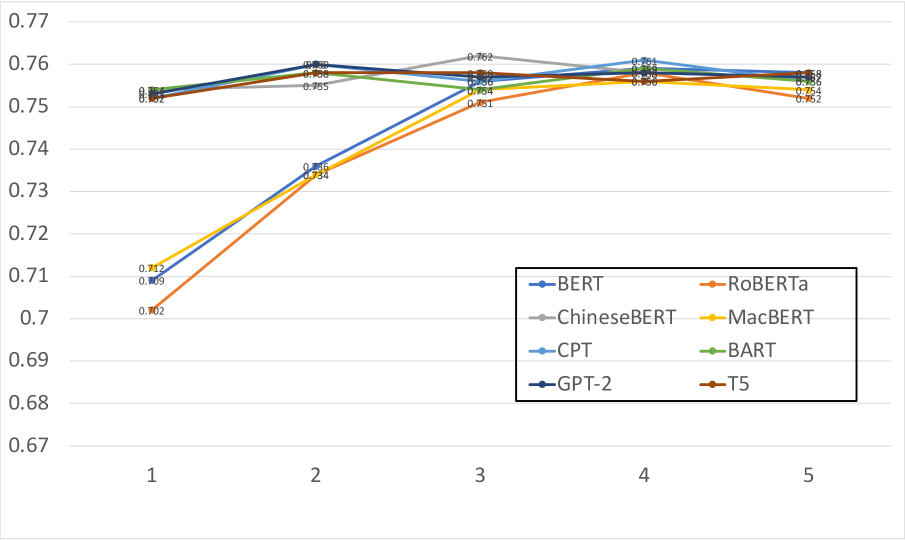}}
\caption{Results for each model in the case of 1-5 layers of MLP. }\label{fig:mlplayers}
\end{figure}

\end{document}